\documentclass{article}

\usepackage{arxiv}

\usepackage[utf8]{inputenc} 
\usepackage[T1]{fontenc}    
\usepackage{hyperref}       
\usepackage{url}            
\usepackage{booktabs}       
\usepackage{amsfonts}       
\usepackage{nicefrac}       
\usepackage{microtype}      
\usepackage{lipsum}
\usepackage{graphicx}
\usepackage{siunitx}
\usepackage{subcaption}
\usepackage[title]{appendix}
\usepackage{float}

\newcommand\blfootnote[1]{%
  \begingroup
  \renewcommand\thefootnote{}\footnote{#1}%
  \addtocounter{footnote}{-1}%
  \endgroup
}

\title{Tracking translation invariance in CNNs}

\author{
  Johannes C.~Myburgh, Coenraad~Mouton, Marelie H.~Davel\\
  Multilingual Speech Technologies, North-West University, South Africa; and CAIR, South Africa.\\
  \texttt{christiaanmyburgh01@gmail.com, moutoncoenraad@gmail.com, marelie.davel@nwu.ac.za} \\
}

\begin{document}
\maketitle
\blfootnote{The final authenticated publication is available online at \url{https://link.springer.com/chapter/10.1007/978-3-030-66151-9_18}}

\begin{abstract}
Although Convolutional Neural Networks (CNNs) are widely used, their translation invariance (ability to deal with translated inputs) is still subject to some controversy. We explore this question using translation-sensitivity maps to quantify how sensitive a standard CNN is to a translated input. We propose the use of Cosine Similarity as sensitivity metric over Euclidean Distance, and discuss the importance of restricting the dimensionality of either of these metrics when comparing architectures. Our main focus is to investigate the effect of different architectural components of a standard CNN on that network’s sensitivity to translation. By varying convolutional kernel sizes and amounts of zero padding, we control the size of the feature maps produced, allowing us to quantify the extent to which these elements influence translation invariance. We also measure translation invariance at different locations within the CNN to determine the extent to which convolutional and fully connected layers, respectively, contribute to the translation invariance of a CNN as a whole. Our analysis indicates that both convolutional kernel size and feature map size have a systematic influence on translation invariance. We also see that convolutional layers contribute less than expected to translation invariance, when not specifically forced to do so.

\end{abstract}


\section{Introduction}
With the impressive performance of Convolutional Neural Networks (CNNs) in object classification~\cite{GradientLearning,CNNAccuracy}, they have become the go-to option for most modern computer vision tasks. Due to their popularity, many different architectural variations of CNNs~\cite{GoogleLeNet,DenseNet,ResNet} have arisen in the past few years that excel at specific tasks. One of the reasons for their rise in popularity is their capability to deal with translated input features. It is widely believed that CNNs are capable of learning translation-invariant representations, but the mechanism behind this translation invariance is still poorly understood. In this study we omit complex variations of the CNN architecture and aim to explore translation invariance in standard CNNs. 

Our goal is to investigate how the various components of a CNN influence and contribute to translation invariance, with a focus on convolutional kernel size, feature map size, convolutional layers and fully connected layers. We achieve this goal by using a translation-sensitivity metric introduced by~\cite{Kauderer} to quantify translation invariance. By investigating CNNs with different convolutional kernel sizes, while implementing zero-padding to control feature map size, we are able to see how convolutional kernel size influences translation invariance. Removing the zero-padding allows us to see how feature map size influences the translation sensitive of a CNN. We also propose a slight change to the translation sensitivity metric that allows us to measure translation invariance within a CNN, allowing us to determine the extent to which convolutional and fully connected layers, respectively, contribute to the translation invariance of a CNN as a whole. We do our analysis on CNNs trained to fit the MNIST dataset and repeat all experiments on the CIFAR10 dataset. 

\section{Related Work}
When investigating translation invariance, we require a performance metric that measures the magnitude of the effect when a sample is translated in a given direction: both direction and magnitude are important. To address this,  Kauderer-Abrams~\cite{Kauderer} developed translation-sensitivity maps that can be used to visualize and quantify exactly how sensitive a network is to shifted inputs. They then use these translation-sensitivity maps to explore the effects of the number of pooling layers and convolutional kernel size on translation invariance. They find that these architectural choices only have a secondary effect, and identify augmented training data as the biggest influence on translation invariance. 

Due to CNNs containing moving kernels, many believe that CNNs are fully translation invariant. Kayman et al. ~\cite{OnTranslationInvariance} challenge this assumption by showing that CNNs have the capability to learn filters that abuse absolute spatial location. Due to the large receptive fields of CNNs, they are able to exploit boundary effects quite far from the image border. In their work they use different forms of padding to remove spatial location encoding which then improves translation invariance. 

In work by Zhang~\cite{Zhang}, he shows that introducing anti-aliasing before sub-sampling, and implementing it correctly, results in higher classification accuracy and better generalization over many architectures on ImageNet~\cite{Imagenet}. Taking a more theoretical approach, Lenc and Vedaldi~\cite{KarelL} explore equivariance, invariance and equivalence in detail. They propose a number of methods to empirically establish these properties and later apply them to a CNN. Their work shows how earlier layers learn to identify general geometrical patterns while deeper layers learn to be more task-specific
\section{Optimization and Architecture}
In this section we discuss the CNN architectures and optimization protocol used in our experiments.

\subsection{General CNN architectural choices}
We investigate the translation invariance of standard CNN architectures, defining such architectures as consisting of multiple sets of convolutional layers and pooling layers, followed by a set of fully connected layers. We do not investigate the effects of dilation, regularization, dropout or skip connections. All networks use Max-Pooling~\cite{Max-Pool} with kernel size and stride of 2, as is commonly used in many popular CNN architectures~\cite{VGG}, and as it works well with the size of our input data. Most popular CNNs contain multiple convolutional layers~\cite{VGG,ResNet,DenseNet}, thus our CNNs all contain 3 convolutional layers, each followed by a Max-Pool layer, connected to a 500 node hidden layer that connects to the output. All architecture-specific details can be found in Appendix A.

\subsection{Datasets}
In our analysis we use the MNIST dataset~\cite{MNIST} containing 28x28 pixel samples of handwritten digits and the CIFAR10 dataset~\cite{CIFAR10} containing 32x32 pixel samples of different class colour images. The MNIST dataset is split into a training set containing 55 000 samples, a validation set containing 5 000 samples and a test set containing 10 000 samples. The CIFAR10 dataset is split into a training set containing 45 000 samples, a validation set containing 5 000 samples and a test set containing 10 000 samples. To be able to generate translation-sensitivity maps without loss of features, all samples are zero-padded with a 6-pixel border.

\subsection{Network Optimization}
All networks are initialized with 
He initialization~\cite{HEInit} using 3 different seeds. Adam is used to optimize Cross-Entropy Loss with a batch size of 128. Four initial learning rates are used: when the best performing learning rate is found at the edge of the learning rate sweep, the learning rate is varied by 0.001 outside the sweep range to ensure that only fully optimized networks are used to generate results. All networks are trained to near-perfect train accuracy ($>99\%$) and are optimized on validation accuracy. All results shown are averaged over 3 seeds.

\section{Translation sensitivity quantification}
In this section we define translation invariance and discuss the sensitivity metric we use to quantify translation invariance. 

\subsection{Translation invariance}
For a system to be completely translation invariant, its output must not be influenced by any translation of the input. The output of a translation-invariant system must thus remain identical for translated and untranslated inputs. Although translation-invariance is a desirable quality for most image classification systems, it is seldom achieved in practice. Knowing that complete translation-invariance is near impossible for standard CNN architectures, we redefine the term “translation-invariance” to refer to a system’s sensitivity to translated inputs. This means that a system can be more or less translation-invariant based on the values received from our translation sensitivity quantification metric.

\subsection{Translation-sensitivity maps}
To quantify and measure translation sensitivity, we use translation-sensitivity maps and radial translation-sensitivity functions, as introduced by \cite{Kauderer}, with a slight change to the sensitivity metric.

Translation-sensitivity maps are 2D graphs that consist of multiple pixels. Each pixel has a value that represents the network’s sensitivity to a specific shift in the input. To calculate the values of these pixels, we determine the similarity between two vectors, namely the base output vector and the translated output vector. These vectors are generated by passing an input sample through a network with the output of the final fully connected layer being referred to as either the base output vector or the translated output vector. The base output vector is generated by passing a non-translated input to the network. The translated output vector is generated by passing a translated input image to the network, with the x-axis and y-axis shifts corresponding to the pixel’s location in the translation-sensitivity map. The similarity between these two vectors is then used as the translation-sensitivity metric. If there is a high similarity between the base output vector and the translated output vector, the network is less sensitive to that specific translation. This high similarity is then represented by a brighter pixel in the translation-sensitivity map. To generate a translation-sensitivity map, we calculate this similarity for each sample with 441 different translations (-10px to 10px shift in the x-axis and -10px to 10px shift in the y-axis) and calculate the average translation-sensitivity map over all samples in a class.

\subsection{Cosine Similarity}
In the introductory paper, the Euclidean Distance between the two vectors is used as similarity metric. The outputs from the Euclidean Distance calculation are non-normalized, restricting comparisons at different locations within a network. (Even if two layers have the same dimensions, the activation values at different layers may have different size distributions.) To address this normalization issue, we propose the use of Cosine Similarity (Eq.~\ref{CosineSim}) to calculate the similarity between the two vectors. Cosine Similarity measures the cosine of the angle between two vectors $\vec{a}$ and $\vec{b}$ in a multi-dimensional space, producing a similarity value between 1 (high similarity) and -1 (high dissimilarity).

\begin{equation}
\label{CosineSim}
cos(\theta) = \frac{\vec{a} \cdotp \vec{b}}{\|\vec{a}\|\|\vec{b}\|
}
\end{equation}

To ensure that the Cosine Similarity measurement produces comparable results to the Euclidean Distance measurement, we use the classification accuracy of a network as a baseline sensitivity measurement to compare the two metrics. We generate translation-sensitivity maps, for each MNIST class, using the three metrics and calculate Pearson Correlation Coefficients to see how correlated Cosine Similarity and Euclidean Distance is with classification accuracy.

\begin{figure}
\includegraphics[width=\textwidth]{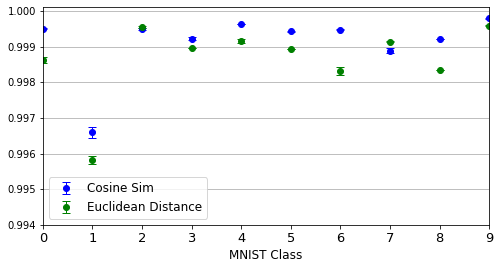}
\caption{Pearson Correlation Coefficients. The Cosine Similarity and Euclidean Distance correlation coefficients with classification accuracy are calculated for each class of the MNIST dataset. These results are generated with the first CNN architecture in Table.~\ref{tab1}. Results are averaged over three seeds. 
} \label{fig:PearsonCoeff}
\end{figure}

Both sensitivity metrics show a very strong positive correlation with classification accuracy (and with each other). From the results in Fig.~\ref{fig:PearsonCoeff}, it seems that Cosine Similarity does provide similar information to Euclidean Distance with the added benefit of the results being normalized, allowing for comparisons across network architecture layers. 

\subsubsection{Restricting Dimensionality }
Although Cosine Similarity has the advantage of producing normalized results, it does not allow for the direct comparison of vectors with different dimensions. When using either Euclidean Distance or Cosine Similarity the dimensions of the datasets greatly influence the calculated metric. As dimension increases, the average Euclidean Distance value tends to increase while the average Cosine Similarity value decreases. This effect produces warped results when comparing either Euclidean Distance or Cosine Similarity values.

\subsection{Radial translation-sensitivity functions}
Radial translation-sensitivity functions are used to compare translation-sensitivity maps. These functions are generated by calculating the radial mean of a translation-sensitivity map at different radii. The output is a set of radial mean values that illustrate how sensitive a network is to the average translation that occurs at a specific radius. To generate these radial translation-sensitivity functions, all samples of the corresponding test set are used to generate an average translation-sensitivity map per class. These translation-sensitivity maps are used to generate radial translation-sensitivity functions for each class. The final radial translations-sensitivity functions shown in our results are the average sensitivity functions for all classes of a network. 

In Fig.~\ref{fig:OutputRadialMeans},~\ref{fig:CIFARShiftedFeatureMaps},~\ref{fig:FMSizes},~\ref{fig:FMSizesCIFAR} the lines show the average sensitivity of a CNN to a shift at a specific radius. The y-axis values show translation invariance with the value of 1 indicating perfect translation invariance and 0 indicating poor translation invariance. 

\section{Tracking Invariance}
As previously stated, standard convolutional neural networks consist mainly of convolutional layers, separated by pooling layers, followed by a final set of fully connected layers. The convolutional layers can be seen as encoders that morph and highlight important features within the input. This is achieved by applying convolutional kernels, with weights that are fine-tuned to identify certain features in the input. The fully connected layers then use these encoded inputs to perform certain tasks. In essence, convolutional layers learn to identify certain features and their characteristics within the input and then pass these features to the following layers in a more efficient representation.  These efficient representations are referred to as Feature Maps and their size depends on several variables such as kernel size, stride, pooling, padding and input size. Although standard CNNs have proven to be less sensitive to shifted inputs than the standard Multilayer Perceptron (MLP), recent research~\cite{OnTranslationInvariance} has shown that they are not, in fact, translation invariant.

In this section, we aim to determine the extent to which convolutional and fully connected layers, respectively, contribute to the translation invariance of a CNN as a whole. Since the fully connected layers of a CNN act as the classifier, it is desired that the inputs they receive be unaffected by input shifts. Although we know that complete translation invariance is unlikely with a standard CNN architecture, it is still desired that the convolutional layers compensate for most of the translation in the input since they are more equipped (with moving kernels and spatial awareness) to deal with translation. 

\subsection{Experiment 1: Translation invariance at different locations within a CNN}
In the first experiment, we investigate the effect of convolutional kernel size on translation invariance on the MNIST dataset. We also investigate how sensitive a standard CNN is to translation at two locations within the network. To test translation-sensitivity at the first location, we use the output of the last convolution layer to generate sensitivity maps. This is done to investigate the effect that convolutional layers have on the network’s sensitivity to translation. For the second location, the output of the final fully connected layer is used to generate translation-sensitivity maps allowing us to see for how much translation invariance the fully connected layers are responsible for.

All convolutional kernel sizes are kept constant throughout each convolutional layer of each network, but varied over the three different networks. Zero-padding is used to ensure that all feature maps have the same size regardless of the change in convolutional kernel size. This is done to allow us to compare convolutional layer outputs across the three CNNs. By keeping the size of the feature maps produced by the final convolutional layer the same across the different CNNs (5x5px), we can be assured that changes in dimensionality do not affect the results. The number of channels per convolutional layer is kept constant across all three networks. The output from the final fully connected layers all have a length of 10 and require no modifications to be comparable with each other. 

\begin{figure}[!tbp]
  \centering
  \subfloat[Final convolutional layer output]{\includegraphics[width=0.5\textwidth]{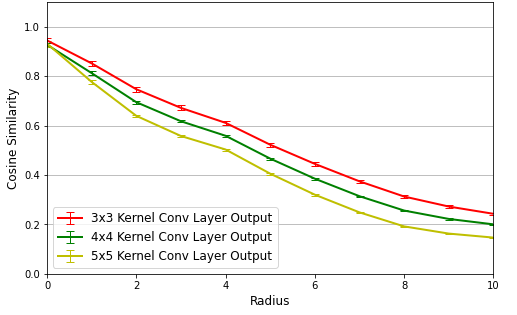}\label{fig:f1}}
  \hfill
  \subfloat[Final fully connected layer output]{\includegraphics[width=0.5\textwidth]{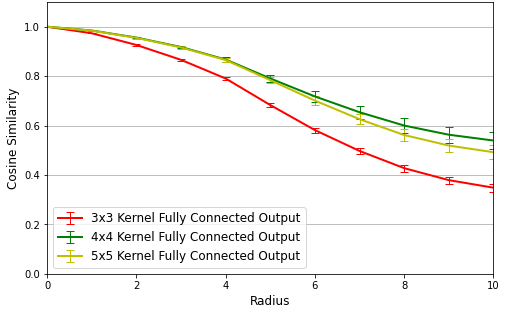}\label{fig:f2}}
  \caption{Radial translation-sensitivity functions generated from both the final convolutional layer output(a) and fully connected layer output(b) on MNIST. Detailed CNN architectures can be found in Table.\ref{tab1}.   }
\label{fig:OutputRadialMeans}
\end{figure}

In Fig.~\ref{fig:OutputRadialMeans}(a) we see the radial translation-sensitivity functions generated from the outputs of the final convolutional layers of the CNNs. It seems that smaller convolutional kernels produce a feature map that is slightly less sensitive to translated inputs. Although it is expected that the convolutional layers would be responsible for most translation invariance of the network, the fully connected layers drastically change the results. The results in Fig.~\ref{fig:OutputRadialMeans}(b) are generated from the final fully connected layers of the CNNs and show that networks with larger convolutional kernel sizes tend to be more translation-invariant after the fully connected layer.

It is interesting to see that although the convolutional layers reduce the network’s sensitivity to translation, it seems that the fully connected layers reduce the network’s translation-sensitivity even more than the convolutional layers. This is somewhat counter-intuitive as it is expected that the convolutional layers would be able to compensate for the translated inputs far better than the fully connected layers.  

\begin{figure}[!tbp]
  \centering
  \subfloat[Non-Shifted Input]{\includegraphics[width=0.25\textwidth]{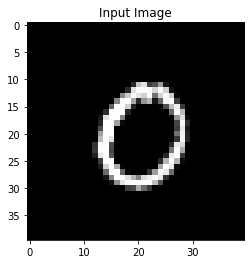}\label{fig:f11}}
  \hfill
  \subfloat[Ch.1 Output]{\includegraphics[width=0.25\textwidth]{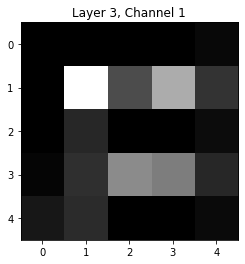}\label{fig:f12}}
  \hfill
  \subfloat[Ch.2 Output]{\includegraphics[width=0.25\textwidth]{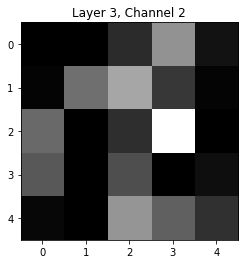}\label{fig:f13}}
  \hfill
  \subfloat[Ch.3 Output]{\includegraphics[width=0.25\textwidth]{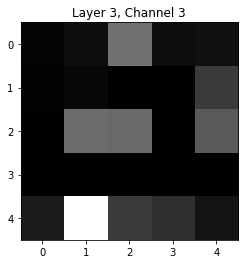}\label{fig:f14}}
  \hfill
  
  \subfloat[Shifted Input]{\includegraphics[width=0.25\textwidth]{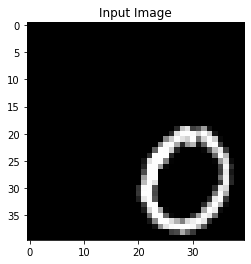}\label{fig:f21}}
  \hfill
  \subfloat[Ch.1 Output]{\includegraphics[width=0.25\textwidth]{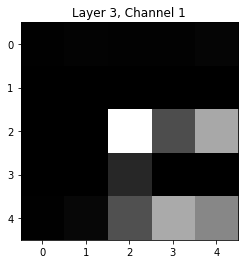}\label{fig:f22}}
  \hfill
  \subfloat[Ch.2 Output]{\includegraphics[width=0.25\textwidth]{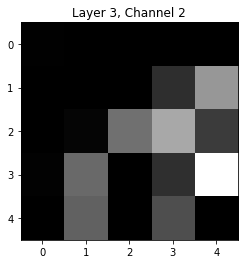}\label{fig:f23}}
  \hfill
  \subfloat[Ch.3 Output]{\includegraphics[width=0.25\textwidth]{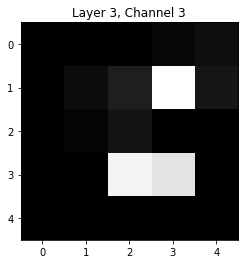}\label{fig:f24}}
  \hfill
  
  \caption{Feature Maps from the final convolutional layer of a CNN given a normal and shifted input sample. These feature maps are arbitrarily chosen to show the presence of translation after three convolutional layers. The translation is present in all feature maps, we only show the first three as is is impossible to show all 30 feature maps.}  
  
\label{fig:ShiftedFeatureMaps}
\end{figure}

To further investigate this effect, we examine the first three output feature maps of the final convolutional layer of the 5x5 kernel size CNN (used in Experiment 1) given a normal and shifted input. The results in Fig.~\ref{fig:ShiftedFeatureMaps} show a normal and shifted input sample and the feature maps produced by the final convolutional layer. One can see that the convolutional layers have highlighted and morphed the input features, but the input shift is still somewhat present in the convolutional layer output feature maps, now forcing the fully connected layers to compensate for the shift. 

One hypothesis is that, since these experiments were performed on a very simple task (MNIST), the convolutional layers were never forced to learn a better encoding scheme, since the fully connected layers were easily able to memorise all of the transformed samples. This would explain the results in Fig.~\ref{fig:OutputRadialMeans}, and why it seems that the fully connected layers contribute more to translation invariance than the convolutional layers. 
One way to test this hypothesis would be to limit the size of the convolutional layer output feature maps, limiting the number of output pixels available to the convolutional layers, and forcing them to learn a more effective encoding scheme. We explore this further in Section 6. 

\begin{figure}[!tbp]
  \centering
  \subfloat[Final convolutional layer output]{\includegraphics[width=0.5\textwidth]{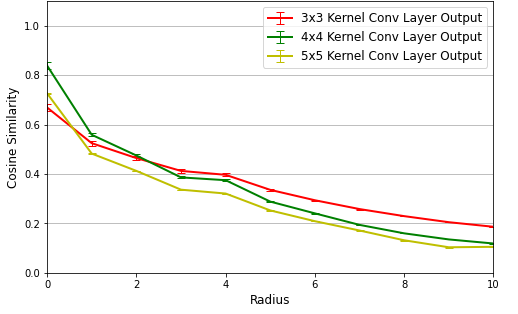}\label{fig:f1}}
  \hfill
  \subfloat[Final fully connected layer output]{\includegraphics[width=0.5\textwidth]{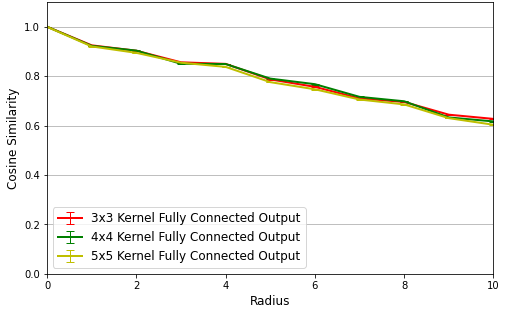}\label{fig:f2}}
  \caption{Radial translation-sensitivity functions generated from both the final convolutional layer output(a) and fully connected layer output(b) on CIFAR10. Detailed CNN architectures can be found in Table.~\ref{tab2}}
\label{fig:CIFARShiftedFeatureMaps}
\end{figure}

Another way to shed light on this observation, is to repeat the experiment on a more complex dataset such as CIFAR10. We increase the number of channels per layer to allow the models to fit the CIFAR10 dataset. In Fig.~\ref{fig:CIFARShiftedFeatureMaps}(a) we see the radial translation-sensitivity functions generated from the final convolutional layer outputs that show a similar trend as in Fig.~\ref{fig:OutputRadialMeans}(a) where smaller convolutional kernels produce less sensitive feature maps. Interestingly, in Fig.~\ref{fig:CIFARShiftedFeatureMaps}(b), the radial translation-sensitivity functions generated from the final fully connected layers, all the outputs seem to have the same level of translation invariance regardless of convolutional kernel size. Although all three networks seem to have the same sensitivity to translation, it is surprising to see how much of an influence the fully connected layers have on the total translation invariance of the CNNs. The inter-class samples in CIFAR10 are less homogeneous than the samples in MNIST, explaining the high translation invariance observed in Fig.~\ref{fig:CIFARShiftedFeatureMaps}(b).

\section{Feature map size and translation invariance}
Having explored two effects of convolutional kernel size on translation invariance when using zero-padding to control feature map output size, we now omit zero-padding to analyse how varying feature map size in conjunction with convolutional kernel size affects translation invariance.

\subsection{Experiment 2: Varying convolutional kernel sizes without zero-padding}
In this experiment we investigate how changes in feature map size due to convolutional kernel size influence translation invariance on MNIST. All convolutional kernel sizes are kept constant throughout each convolutional layer of each network, but varied over the different networks. No zero-padding is used, allowing for reduced feature map sizes. Over all three networks, varying amounts of channels are added to keep the number of effective nodes (kernel size \(\times\) number of channels) per layer comparable.

\begin{figure}
\includegraphics[width=\textwidth]{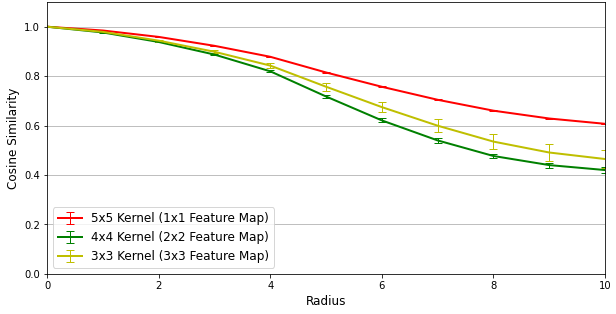}
\caption{Radial Translation-sensitivity functions of CNNs with different convolutional kernel sizes on MNIST. These results are generated from the fully connected layer outputs of the CNNs since varying feature map sizes do not allow for direct comparisons of feature maps generated by final convolutional layers. Detailed CNN architectures can be found in Table~\ref{tab3}.   
} \label{fig:FMSizes}
\end{figure}

Results are shown in  Fig.~\ref{fig:FMSizes}. These results show the translation-sensitivity functions calculated from the final fully connected layer outputs of the CNNs. It is clear that  there is a large increase in translation invariance when the feature map reaches a size of 1x1 pixel. Since there is no room for movement in a 1-pixel feature map, it seems that the convolutional layers are forced to better deal with input translations. Although it is clear that fully connected layers still add translation invariance, it does seem that reducing the feature map output size does not have significant effect on translation invariance.

With the benefit of less translation sensitive networks, why not reduce all convolutional layer output feature maps to 1x1 pixel? This is indeed possible, as is the case with fully convolutional networks. However, these networks limit capacity in a way that additional fully connected layers do not. Neural Networks trained for a classification task require adequate capacity to be able to learn the characteristics and features of the different classes. Small map sizes are possible for tasks such as MNIST, but reducing feature map size can become a bottleneck in larger networks trained for more demanding classification tasks. 

\begin{figure}[]
\includegraphics[width=\textwidth]{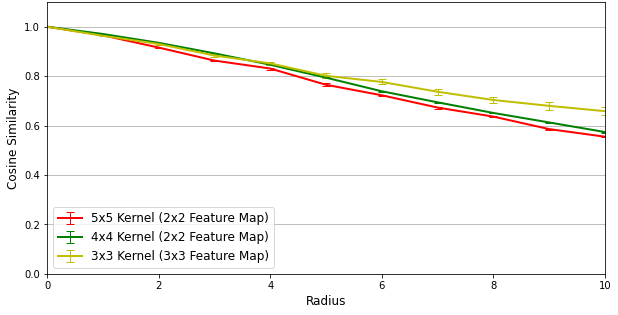}
\caption{Radial Translation-sensitivity functions of CNNs with different convolutional kernel sizes on CIFAR10. These results are generated from the fully connected layer outputs of the CNNs since varying feature map sizes do not allow for direct comparisons of feature maps generated by final convolutional layers. Detailed CNN architectures can be found in Table.~\ref{tab4}   
} \label{fig:FMSizesCIFAR}
\end{figure}

We repeat Experiment 2 on the CIFAR10 dataset to see how feature map size would affect a CNN trained on a more complex dataset with less homogeneous samples. The architectures we use are very similar to the architectures used on MNIST, except we add more channels to give the CNNs sufficient capacity to fit the more complex CIFAR10 dataset.\\ 

The results in Fig.~\ref{fig:FMSizesCIFAR}, the radial translation-sensitivity functions generated from the fully connected layer outputs, show a similar pattern as the results in Fig.~\ref{fig:FMSizes}. It seems that reducing feature map size has little to no influence on translation invariance, especially in a CNN trained on a more complex dataset that forces the network to be more translation invariant during training.  

\section{Conclusion}
In this paper we use translation-sensitivity maps to analyse how different components of a standard CNN affect the network’s translation invariance. We train several standard CNNs on the MNIST and CIFAR10 datasets. We propose a slight change to the similarity metric and prove that it produces comparable results with the added benefit of normalizing results. Specifically, we focus on convolutional kernel size and find that smaller kernels tend to produce feature maps that are less sensitive to translated inputs. We also study how convolutional and fully connected layers affect translation invariance and find that although convolutional layers contribute, it seems that fully connected layers are responsible for the majority of translation invariance in a standard CNN. In our study we also vary feature map size and find that it has little effect on translation sensitivity.   

\newpage
\bibliographystyle{unsrt}  
\bibliography{references} 

\begin{thebibliography}{10}

\bibitem{GradientLearning}
Y.~{Lecun}, L.~{Bottou}, Y.~{Bengio}, and P.~{Haffner}.
\newblock Gradient-based learning applied to document recognition.
\newblock {\em Proceedings of the IEEE}, 86(11):2278--2324, 1998.

\bibitem{CNNAccuracy}
Alex Krizhevsky, Ilya Sutskever, and Geoffrey Hinton.
\newblock Imagenet classification with deep convolutional neural networks.
\newblock {\em Neural Information Processing Systems}, 25, 01 2012.

\bibitem{GoogleLeNet}
Christian Szegedy, Wei Liu, Yangqing Jia, Pierre Sermanet, Scott~E. Reed,
  Dragomir Anguelov, Dumitru Erhan, Vincent Vanhoucke, and Andrew Rabinovich.
\newblock Going deeper with convolutions.
\newblock {\em CoRR}, abs/1409.4842, 2014.

\bibitem{DenseNet}
Gao Huang, Zhuang Liu, and Kilian~Q. Weinberger.
\newblock Densely connected convolutional networks.
\newblock {\em CoRR}, abs/1608.06993, 2016.

\bibitem{ResNet}
Kaiming He, Xiangyu Zhang, Shaoqing Ren, and Jian Sun.
\newblock Deep residual learning for image recognition.
\newblock {\em CoRR}, abs/1512.03385, 2015.

\bibitem{Kauderer}
Eric Kauderer{-}Abrams.
\newblock Quantifying translation-invariance in convolutional neural networks.
\newblock {\em CoRR}, abs/1801.01450, 2018.

\bibitem{OnTranslationInvariance}
Osman~Semih Kayhan and Jan C.~van Gemert.
\newblock On translation invariance in cnns: Convolutional layers can exploit
  absolute spatial location.
\newblock In {\em Proceedings of the IEEE/CVF Conference on Computer Vision and
  Pattern Recognition (CVPR)}, June 2020.

\bibitem{Zhang}
Richard Zhang.
\newblock Making convolutional networks shift-invariant again.
\newblock {\em CoRR}, abs/1904.11486, 2019.

\bibitem{Imagenet}
J.~Deng, W.~Dong, R.~Socher, L.-J. Li, K.~Li, and L.~Fei-Fei.
\newblock {ImageNet: A Large-Scale Hierarchical Image Database}.
\newblock In {\em CVPR09}, 2009.

\bibitem{KarelL}
Karel Lenc and Andrea Vedaldi.
\newblock Understanding image representations by measuring their equivariance
  and equivalence.
\newblock {\em CoRR}, abs/1411.5908, 2014.

\bibitem{Max-Pool}
Dominik Scherer, Andreas Müller, and Sven Behnke.
\newblock Evaluation of pooling operations in convolutional architectures for
  object recognition.
\newblock pages 92--101, 01 2010.

\bibitem{VGG}
Karen Simonyan and Andrew Zisserman.
\newblock Very deep convolutional networks for large-scale image recognition.
\newblock {\em CoRR}, abs/1409.1556, 2014.

\bibitem{MNIST}
Yann LeCun and Corinna Cortes.
\newblock {MNIST} handwritten digit database.
\newblock 2010.

\bibitem{CIFAR10}
Alex Krizhevsky, Vinod Nair, and Geoffrey Hinton.
\newblock Cifar-10 (canadian institute for advanced research).

\bibitem{HEInit}
Kaiming He, Xiangyu Zhang, Shaoqing Ren, and Jian Sun.
\newblock Delving deep into rectifiers: Surpassing human-level performance on
  imagenet classification.
\newblock {\em CoRR}, abs/1502.01852, 2015.

\end{thebibliography}
\newpage
\newpage

\begin{appendices}
\section{Appendix A: Network Architectures}

Here we show all network architectures and performances of the CNNs used in our experiments.

\begin{table}[H]
\centering
\caption{MNIST Architectures used in Experiment 1.}\label{tab1}
\begin{tabular}{|l|l|l|l|l|l|l|}
\hline
{\textbf Layer} & {\textbf Kernel Size} & {\textbf Stride} & {\textbf Padding} & {\textbf Output Size} & {\textbf Channels} & {\textbf Activation} \\
\hline
CNN1      & 5,4,3       & 1      & 2,2,1   & 40          & 10       & ReLU                \\
Max-Pool1 & 2           & 2      & 0       & 20          & -        &                     \\
CNN2      & 5,4,3       & 1      & 2,2,1   & 20          & 20       & ReLU                \\
Max-Pool2 & 2           & 2      & 0       & 10          & -        &                     \\
CNN3      & 5,4,3       & 1      & 2,1,1   & 10          & 30       & ReLU                \\
Max-Pool3 & 2           & 2      & 0       & 5           & -        &                     \\
FC        & -           & -      & -       & 500         & -        & ReLU                \\
Out       & -           & -      & -       & 10          & -        & Softmax            \\
\hline
\end{tabular}
\end{table}

All networks in Table.\ref{tab1} achieve a minimum training accuracy of 100\%, validation accuracy of 99.16\% and test accuracy of 99.19\%.

\begin{table}[H]
\centering
\caption{CIFAR10 Architectures used in Experiment 1.}\label{tab2}
\begin{tabular}{|l|l|l|l|l|l|l|}
\hline
{\textbf Layer} & {\textbf Kernel Size} & {\textbf Stride} & {\textbf Padding} & {\textbf Output Size} & {\textbf Channels} & {\textbf Activation} \\
\hline
CNN1      & 5,4,3       & 1      & 2,2,1   & 44          & 50       & ReLU                \\
Max-Pool1 & 2           & 2      & 0       & 22          & -        &                     \\
CNN2      & 5,4,3       & 1      & 2,2,1   & 22          & 100       & ReLU                \\
Max-Pool2 & 2           & 2      & 0       & 11          & -        &                     \\
CNN3      & 5,4,3       & 1      & 2,1,1   & 11          & 150       & ReLU                \\
Max-Pool3 & 2           & 2      & 0       & 5           & -        &                     \\
FC        & -           & -      & -       & 500         & -        & ReLU                \\
Out       & -           & -      & -       & 10          & -        & Softmax            \\
\hline
\end{tabular}
\end{table}

All networks in Table.\ref{tab2} achieve a minimum training accuracy of 99.89\%, validation accuracy of 74.45\% and test accuracy of 74.67\%.

\begin{table}[H]
\centering
\caption{MNIST Architectures used in Experiment 2.}\label{tab3}
\begin{tabular}{|l|l|l|l|l|l|l|}
\hline
{\textbf Layer} & {\textbf Kernel Size} & {\textbf Stride} & {\textbf Padding} & {\textbf Output Size} & {\textbf Channels} & {\textbf Activation} \\
\hline
CNN1      & 5,4,3       & 1      & 0       & 36,37,38          & 10,16,28       & ReLU                \\
Max-Pool1 & 2           & 2      & 0       & 18,18,19          & -        &                     \\
CNN2      & 5,4,3       & 1      & 0       & 14,15,17          & 20,31,56       & ReLU                \\
Max-Pool2 & 2           & 2      & 0       & 7,7,8          & -        &                     \\
CNN3      & 5,4,3       & 1      & 0       & 3,4,6          & 30,47,83       & ReLU                \\
Max-Pool3 & 2           & 2      & 0       & 1,2,3           & -        &                     \\
FC        & -           & -      & -       & 500         & -        & ReLU                \\
Out       & -           & -      & -       & 10          & -        & Softmax            \\
\hline
\end{tabular}
\end{table}

All networks in Table.\ref{tab3} achieve a minimum training accuracy of 99.99\%, validation accuracy of 99.18\% and test accuracy of 99.30\%.

\begin{table}[H]
\centering
\caption{CIFAR10 Architectures used in Experiment 2.}\label{tab4}
\begin{tabular}{|l|l|l|l|l|l|l|}
\hline
{\textbf Layer} & {\textbf Kernel Size} & {\textbf Stride} & {\textbf Padding} & {\textbf Output Size} & {\textbf Channels} & {\textbf Activation} \\
\hline
CNN1      & 5,4,3       & 1      & 0       & 40,41,42          & 50,78,139       & ReLU                \\
Max-Pool1 & 2           & 2      & 0       & 20,20,21          & -        &                     \\
CNN2      & 5,4,3       & 1      & 0       & 16,17,19          & 100,156,278       & ReLU                \\
Max-Pool2 & 2           & 2      & 0       & 8,8,9          & -        &                     \\
CNN3      & 5,4,3       & 1      & 0       &4,5,7          & 150,234,416       & ReLU                \\
Max-Pool3 & 2           & 2      & 0       & 2,2,3           & -        &                     \\
FC        & -           & -      & -       & 500         & -        & ReLU                \\
Out       & -           & -      & -       & 10          & -        & Softmax            \\
\hline
\end{tabular}
\end{table}

All networks in Table.\ref{tab4} achieve a minimum training accuracy of 99.93\%, validation accuracy of 74.31\% and test accuracy of 74.40\%.

\end{appendices}

\end{document}